\documentclass[letterpaper, 10 pt, conference]{ieeeconf}  

\IEEEoverridecommandlockouts

\usepackage[utf8]{inputenc}
\usepackage[T1]{fontenc}
\usepackage{cite}
\usepackage{amsmath,amssymb,amsfonts}
\usepackage{graphicx}
\usepackage{textcomp}
\usepackage{xcolor}
\usepackage{lipsum}
\usepackage{booktabs}
\usepackage{subcaption}
\usepackage[hyphens]{url}
\usepackage[belowskip=-4pt,aboveskip=1.5pt]{caption}
\usepackage{algorithm}
\usepackage{algpseudocode}

\IEEEoverridecommandlockouts                              

\title{\LARGE \bf
Towards Interaction State Recognition in Sensorized Gecko-Inspired Adhesive Interfaces for Orbital Servicing}

\author{Eduart Abdullahu$^{1}$, Francesco Grella$^{1}$, Carol Martinez$^{1}$
\thanks{$^{1}$Space Robotics Research Group (SpaceR) , Interdisciplinary Centre for Security, Reliability and Trust (SnT), University of Luxembourg
     {\tt\small  \{\tt\small eduart.abdullahu francesco.grella,
\tt\small  carol.martinezluna\}@uni.lu}}}

\begin{document}

\maketitle
\thispagestyle{empty}
\pagestyle{empty}


\begin{abstract}

Reliable contact establishment remains a critical challenge for autonomous on-orbit servicing (OOS), particularly when interacting with non-cooperative targets. Gecko-inspired dry adhesives provide a lightweight, passive, and residue-free solution for robotic capture in space, yet current systems operate in open loop, without the ability to estimate their own contact or adhesion state during interaction. This paper presents a real-time interaction-awareness framework that enables sensorized gecko-inspired adhesive interfaces to estimate their contact state throughout the adhesion lifecycle using embedded capacitive sensing. Five discrete interaction states (idle, proximity, contact, engaged, and disengaged) are classified using experimentally collected and augmented sensor data, enabling a comparative evaluation of six lightweight machine learning models suitable for embedded deployment. Among the evaluated models, Random Forest achieved the highest offline classification accuracy of 95.65\%. For online deployment, combining Random Forest with a rule-based state machine improved accuracy from 78.02\% to 93.14\%, achieving a macro F1-score of 0.9343 while maintaining a 53 ms inference cycle. These results demonstrate that incorporating physics-consistent state constraints significantly improves robustness during real-time inference, representing a step toward interaction-aware capture in future autonomous orbital servicing missions.

\end{abstract}


\section{Introduction}

Since the launch of Sputnik-1, Earth orbits have become increasingly congested \cite{Ben-Larbi2022}. The planned deployment of megaconstellations is expected to intensify this congestion and increase the risk of collisions and debris generation \cite{SpaceDebris}. Addressing this challenge and ensuring long-term orbital sustainability require advanced debris-removal and on-orbit servicing (OOS) systems capable of establishing safe and reliable initial contact with non-cooperative targets.

Gecko-inspired dry adhesives offer a promising solution for first contact with space structures. Unlike conventional mechanisms, they require no suction or mechanical interlocking, can attach and detach from unprepared surfaces without leaving residue, and can operate in vacuum environments ~\cite{Ben-Larbi2022} with demonstrated performance under space-relevant thermal conditions~\cite{Emj2025}. Inspired by the hierarchical setae structure of gecko feet, these materials rely on van der Waals interactions, enabling adhesion without chemical bonding, contamination, or consumable materials. In particular, wedge-type architectures require only low normal preload combined with controlled shear displacement, minimising reaction forces during contact and making them well-suited for repeated attachment cycles when interacting with free-floating targets.

\begin{figure}[!t]
  \centering
  \includegraphics[width=1.0\columnwidth]{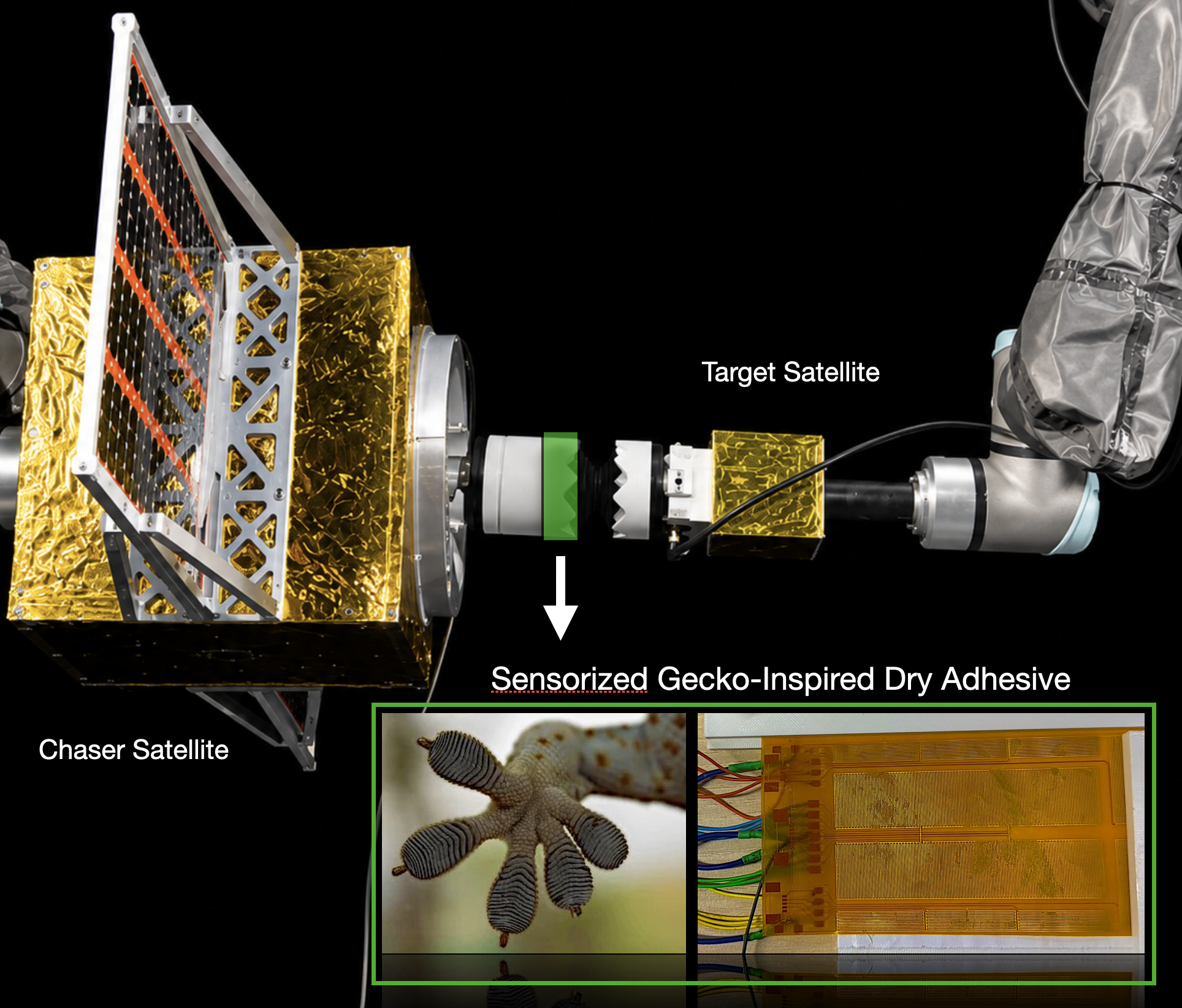}
  \caption{Potential docking scenario. Sensorized gecko-inspired interfaces can provide interaction-state feedback to support autonomous decision-making in future OOS missions.}
  \label{fig:summary_img}

\end{figure}

Despite successful space demonstrations \cite{Jiang2015}\cite{Jiang2017}, reliable use of gecko adhesives remains challenging because of sensitivity to contact trajectory, misalignment, and variability in applied normal and shear loads \cite{Chen2022}. Most systems are open-loop and lack awareness of contact state, thereby limiting autonomy and making recovery from partial engagement or premature release difficult. Moving from passive interfaces toward intelligent adhesive systems, therefore, requires real-time interpretation of the contact interface during interaction.




Integrating sensing into gecko-inspired adhesives can address this gap. Capacitive sensing with embedded flexible electrodes enables mechanically transparent measurement without affecting adhesion \cite{Hashizume2019}. However, the resulting signals are complex and capture multiple overlapping interaction phenomena (proximity effects, initial contact, conformational engagement, and shear loading), requiring data-driven methods to extract meaningful information.

This paper presents an initial step toward intelligent gecko-inspired interfaces by enabling real-time interaction-state estimation in sensorized adhesives. As shown in Fig.~\ref{fig:summary_img}, such awareness of the interaction can provide feedback on the state of the adhesive interface during capture, supporting higher-level decision-making in future autonomous OOS missions. Using a wedge-shaped gecko-inspired adhesive equipped with an embedded interdigitated capacitive sensor developed by the Stanford Biomimetics and Dexterous Manipulation Lab (BDML) \cite{Chen2022}\cite{Emj2025}, we collect experimental data throughout the adhesion lifecycle, including approach, contact, engagement, and release. We apply signal processing and feature extraction methods to characterize the evolving adhesive interaction, then evaluate six lightweight machine learning models for real-time contact-state classification. To improve robustness during dynamic operation, classifier predictions are combined with a state machine that enforces physically consistent interaction transitions.


The main contribution of this work is a real-time interaction-state recognition framework for sensorized gecko-inspired adhesive interfaces, combining lightweight machine learning with physics-informed state-transition constraints for robust online deployment. Experimental validation demonstrates that incorporating these constraints improves online inference robustness, increasing classification accuracy from 78.02\% to 93.14\% during real-time operation, enabling reliable recognition of the adhesive interaction throughout the contact lifecycle.


To the best of our knowledge, this is the first work enabling real-time interaction awareness in sensorized gecko-inspired adhesive interfaces for autonomous space robotic applications, moving adhesive-based capture systems from passive mechanisms toward interaction-aware autonomous operation.



\section{Related Work}

Gecko-inspired dry adhesives have emerged as a promising technology for robotic capture and manipulation in space due to their lightweight, passive, and reusable adhesion mechanism. Early work on ON-OFF adhesive grippers demonstrated the feasibility of reversible, residue-free attachment for orbital applications \cite{Aaron2013}. Subsequent developments have focused on improving the reliability and controllability of adhesive contact \cite{Jiang2017}. A major milestone was the demonstration of gecko-inspired adhesives on the International Space Station using NASA’s Astrobee platform, which validated their performance in true microgravity and highlighted operational challenges, including alignment sensitivity, impact rebound, incomplete contact formation, and variability in normal and shear loading \cite{Chen2022}. Related efforts, such as mushroom-type gecko inspired docking mechanisms \cite{trentlage2016adaptable}, compliant debris capture systems \cite{Max2023}, and commercial technologies such as Kall Morris Inc.’s REACCH platform \cite{kallmorris2025reacch}, further demonstrate the potential of adhesive-based capture and docking systems for orbital applications. 

\begin{figure*}[t]
  \centering
  \includegraphics[width=0.95\textwidth]{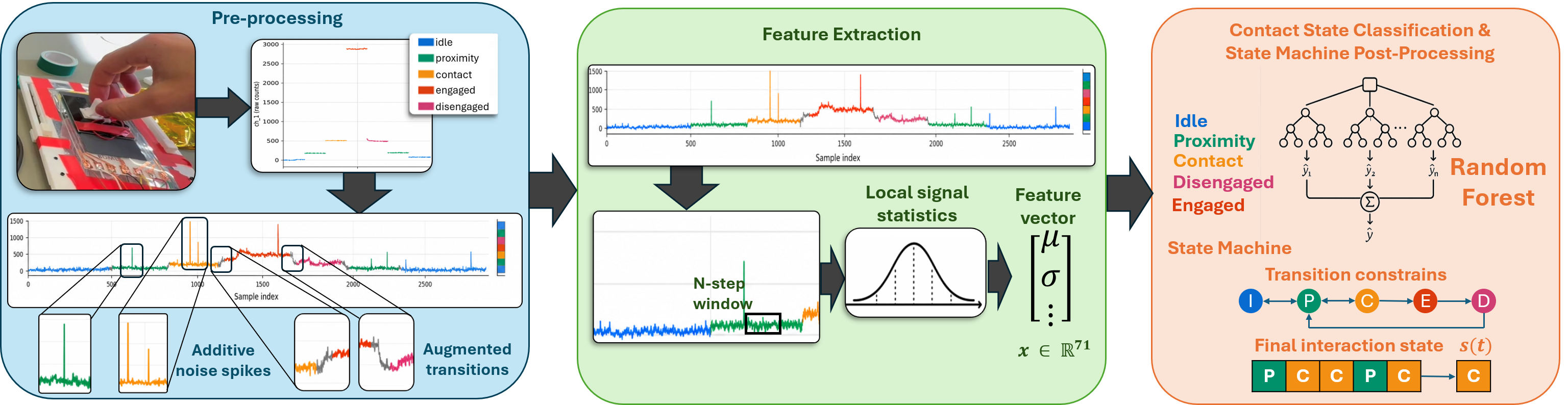}
  \caption{Interaction-state recognition pipeline. Raw steady-state measurements are augmented with synthetic transitions, noise, and sporadic spikes. Statistical features are extracted using an \(N\)-step sliding window and classified by the Random Forest, followed by state-machine post-processing to obtain the final interaction-state estimate.}
  \label{fig:method_pipeline}
\end{figure*}

To improve the reliability and controllability of adhesive interaction, recent research has increasingly focused on integrating sensing directly into adhesive interfaces. In~\cite{liu2025}, a thin-film pressure sensor enabled the detection of adhesion, slip, and detachment states, while piezoelectric sensing based on Lamb wave attenuation has been used to correlate contact area with adhesion behavior \cite{Huh2018}. Capacitive sensing with interdigitated electrodes enables estimation of contact and adhesion state through changes in electric field distribution \cite{Hashizume2019}. More recent designs incorporate compliant adhesives with embedded electrodes for simultaneous contact detection and mechanical adaptability \cite{Li2025}, as well as resistive-capacitive electrode arrays that detect adhesion and deformation, particularly on rough surfaces \cite{Shao2025}. Vision-based tactile sensing has also been explored, using internal imaging to infer contact distribution and shear forces beneath the adhesive layer \cite{Pang2021}. Together, these works demonstrate that adhesive interfaces can be instrumented using a variety of sensing modalities while preserving their mechanical functionality.

However, existing work primarily focuses on sensing and signal acquisition, with limited attention devoted to interpreting these measurements as high-level interaction states that can directly support autonomous robotic decision-making. In particular, real-time inference of the complete contact progression, from approach and initial contact to adhesive engagement and release, remains largely unexplored for space robotic applications. This work addresses this gap by investigating the feasibility of a real-time interaction-state recognition framework capable of translating continuous sensor measurements into discrete, physically meaningful states during online operation.

\section{Interaction State Recognition}
Capacitive sensors embedded within gecko-inspired adhesive interfaces produce continuous time-series signals whose amplitude and temporal evolution reflect changes in the adhesive-surface interaction. Translating these
signals into discrete, physically meaningful states for online monitoring presents
several challenges: inter-session baseline drift, variability in the
applied contact conditions, and the limited availability of
densely labelled experimental data. This section presents the proposed processing pipeline, comprising signal pre-processing, data augmentation, feature extraction, machine learning-based classification, and state-machine post-processing. The complete workflow is illustrated in Fig.~\ref{fig:method_pipeline}.


\subsection{Pre-Processing and Data Augmentation}

Raw capacitance measurements exhibit two principal sources of
variability that must be addressed before classification: absolute
baseline drift across experimental sessions, caused by environmental
conditions such as temperature variations, and the limited diversity of experimentally acquired data. To compensate for baseline drift, each acquisition session is normalised by subtracting a reference capacitance measured during an initial resting period of approximately 20~s, prior to data collection. The  compensated signal is expressed as:
\begin{equation}
  \Delta C(t)=C(t)-C_{\mathrm{base}}
  \label{eq:comp}
\end{equation}

where $C(t)$ is the measured capacitance and $C_{\mathrm{base}}$ is the reference value computed before each experiment. This normalisation reduces inter-session variability and is critical for generalisation. Classifiers trained on $\Delta C$ are not exposed to session-specific absolute levels, therefore preventing the classifier from exploiting session-dependent offsets rather than contact-related information.



Obtaining a large number of labelled interaction sequences under controlled conditions is time-consuming and experimentally demanding. To improve dataset diversity, a targeted data augmentation strategy was therefore adopted. Starting from short steady-state recordings for each interaction state, synthetic sequences were generated while preserving the characteristic capacitance level of the original state. Realistic measurement noise and transient disturbances, estimated from experimentally acquired signals, were then introduced to reproduce the variability observed during real operation. This approach increases the size and diversity of the training set while maintaining physically meaningful signal characteristics, thereby improving robustness to measurement noise and minor variations in the interaction process without requiring additional experimental trials.
This synthetic noise injection intentionally amplifies sensor noise and baseline drift, forcing classifiers to learn noise-resilient decision boundaries for dynamic deployment.

\subsection{Feature Extraction}

Reliable interaction-state recognition requires features that capture both the instantaneous contact condition and its short-term temporal evolution. To provide temporal context while maintaining low inference latency, a sliding window of $N=20$ samples is applied to the compensated capacitance signal. This window length was selected as a compromise between temporal resolution and responsiveness for online operation.

For transition-state classification, each window is additionally interpolated to 60 uniformly spaced samples to obtain a smoother higher-resolution representation of the signal evolution near state boundaries. From each window, the statistical and temporal descriptors listed in Table~\ref{tab:features} are extracted, including measures of central tendency, variability, temporal trend, and autocorrelation, together with the interpolated signal samples. These descriptors are concatenated into a 71-dimensional feature vector, $\mathbf{x}\in\mathbb{R}^{71}$, which serves as the input to the machine learning classifiers. The feature set empirically combines central descriptors with waveform interpolation, where intentional redundancy enhances noise robustness while maintaining a compact vector dimensionality to prevent overfitting on constrained datasets.



\begin{table}[!h]
\centering
\vspace*{0.5cm}
\caption{Extracted signal features for classification}
\label{tab:features}
\begin{tabular}{ll}
\hline
\textbf{Feature} & \textbf{Description} \\
\hline
Mean ($\mu$) & Average signal amplitude \\
Median & Robust central tendency estimate \\
Standard deviation ($\sigma$) & Signal variability \\
Variance ($\sigma^2$) & Dispersion magnitude \\
Minimum / Maximum & Signal extrema \\
Range & Peak-to-peak amplitude span \\
Slope & Linear temporal trend \\
Skewness & Distribution asymmetry \\
Autocorrelation ($lag=1$) & Short-term temporal dependence \\
MAD & Robust variability metric \\
60 interpolation points & Dynamic waveform representation \\
\hline
\end{tabular}
\end{table}

\subsection{Interaction State Classification}

The interaction state recognition problem is formulated as a multi-class
classification problem, in which each feature vector is assigned
to one of the five interaction states. Given the compact feature representation, the moderate size of the labelled dataset, and the computational constraints associated with embedded deployment, this work focuses on classical supervised learning methods rather than deep neural networks. Classical models offer low computational complexity, straightforward interpretability, and efficient real-time execution while providing a suitable baseline for evaluating the feasibility of interaction-state recognition.


To determine the most effective algorithm for this task, six classifiers were evaluated: Support Vector Machine (SVM), k-Nearest Neighbours (k-NN), Logistic Regression (LR), Gradient Boosting (GB), Extra Trees (ET), and Random Forest (RF). Hyperparameters for each model were optimised using the Optuna \cite{akiba2019} optimisation framework. Classifier generalisation was assessed using 10-fold stratified cross-validation on the training set, while the final evaluation was performed on a held-out test set comprising 20\% of the data, with class proportions preserved through stratified sampling.

\subsection{State Machine Post-Processing}
To improve the robustness and physical consistency of the predicted interaction states, a lightweight rule-based State Machine (SM) is applied as a post-processing stage within the inference pipeline. The SM combines temporal filtering with physics-informed transition constraints. Temporal filtering suppresses isolated prediction errors, while the transition rules ensure that the estimated states follow the physically admissible progression of the adhesive interaction.

A first-in, first-out (FIFO) buffer stores the five most recent interaction-state labels. At each inference step, the current accepted state is computed as the majority label within the buffer,
\begin{equation}
s(k)=\operatorname{mode}
\left(
q_{k-4},
q_{k-3},
q_{k-2},
q_{k-1},
q_{k}
\right),
\label{eq:sm}
\end{equation}

where $q_k$ represents the state inserted into the queue at time step $k$, and $\text{mode}(\cdot)$ returns the most frequently occurring label. The queue can contain a mix of raw classifier predictions $\hat{s}$ and next physically admissible states $\tilde{s}$, or consist entirely of one of them depending on transition validity. This majority-voting scheme reduces short-duration fluctuations and isolated misclassifications in the classifier output.


The state-machine update is performed as follows:
\begin{enumerate}
    \item Compute the majority vote of the current FIFO buffer. This gives the previously accepted state $s_{\mathrm{prev}}$.
    \item Obtain the new classifier prediction, $\hat{s}_k$.
    \item Check whether the transition from $s_{\mathrm{prev}}$ to $\hat{s}_k$ is permitted according to Table~\ref{tab:transition_matrix}.
    \item If the transition is valid, insert $\hat{s}_k$ into the FIFO buffer.
    \item If the transition is invalid, replace $\hat{s}_k$ with the next physically admissible intermediate state, denoted as $\tilde{s}_k$, and insert this replacement state into the FIFO buffer.
    \item Recompute the majority vote of the updated buffer. This value is reported as the final output of the state machine.
\end{enumerate}

The transition matrix in Table~\ref{tab:transition_matrix} defines the admissible state changes used in Step~3. Rows correspond to the previously accepted state $s_{\mathrm{prev}}$, while columns correspond to the candidate classifier prediction $\hat{s}_k$. A transition marked as \textit{True} is accepted directly. Otherwise, the classifier prediction is considered physically inconsistent with the expected adhesion process and is replaced by an admissible intermediate state $\tilde{s}_k$ before updating the buffer.

For example, a direct transition from \textit{Proximity} to \textit{Disengaged} is rejected because adhesive release must first pass through the \textit{Contact} and \textit{Engaged} states. In this case, the state machine inserts the next admissible intermediate state instead of the invalid prediction before recomputing the majority vote.

Since the FIFO buffer is initialized with the \textit{Idle} state, a new interaction state is reported once three consistent predictions become the majority of the five-sample window. Therefore, the temporal filtering introduces a maximum delay of approximately three samples while improving the stability and physical consistency of the predicted interaction states. The proposed SM has negligible computational overhead and is suitable for real-time deployment.

\begin{table}[!]
\centering
\vspace*{0.25cm}
\caption{Permitted interaction-state transitions enforced by the state machine. Rows indicate the current state and columns the admissible subsequent states.}
\label{tab:transition_matrix}
\resizebox{\columnwidth}{!}{%
\begin{tabular}{lccccc}
\toprule
\textbf{State} & \textbf{Idle} & \textbf{Prox} & \textbf{Contact} & \textbf{Engaged} & \textbf{Disengaged} \\
\midrule
\textbf{Idle} & True & True & - & - & - \\
\textbf{Prox} & True & True & True & - & - \\
\textbf{Contact} & - & True & True & True & - \\
\textbf{Engaged} & - & - & - & True & True \\
\textbf{Disengaged} & - & True & - & - & True \\
\bottomrule
\end{tabular}%
}
\end{table}

\section{Experiments and Results}
\subsection{Experimental Setup}

\begin{figure}[!bh]
  \centering
  \includegraphics[width=\columnwidth]{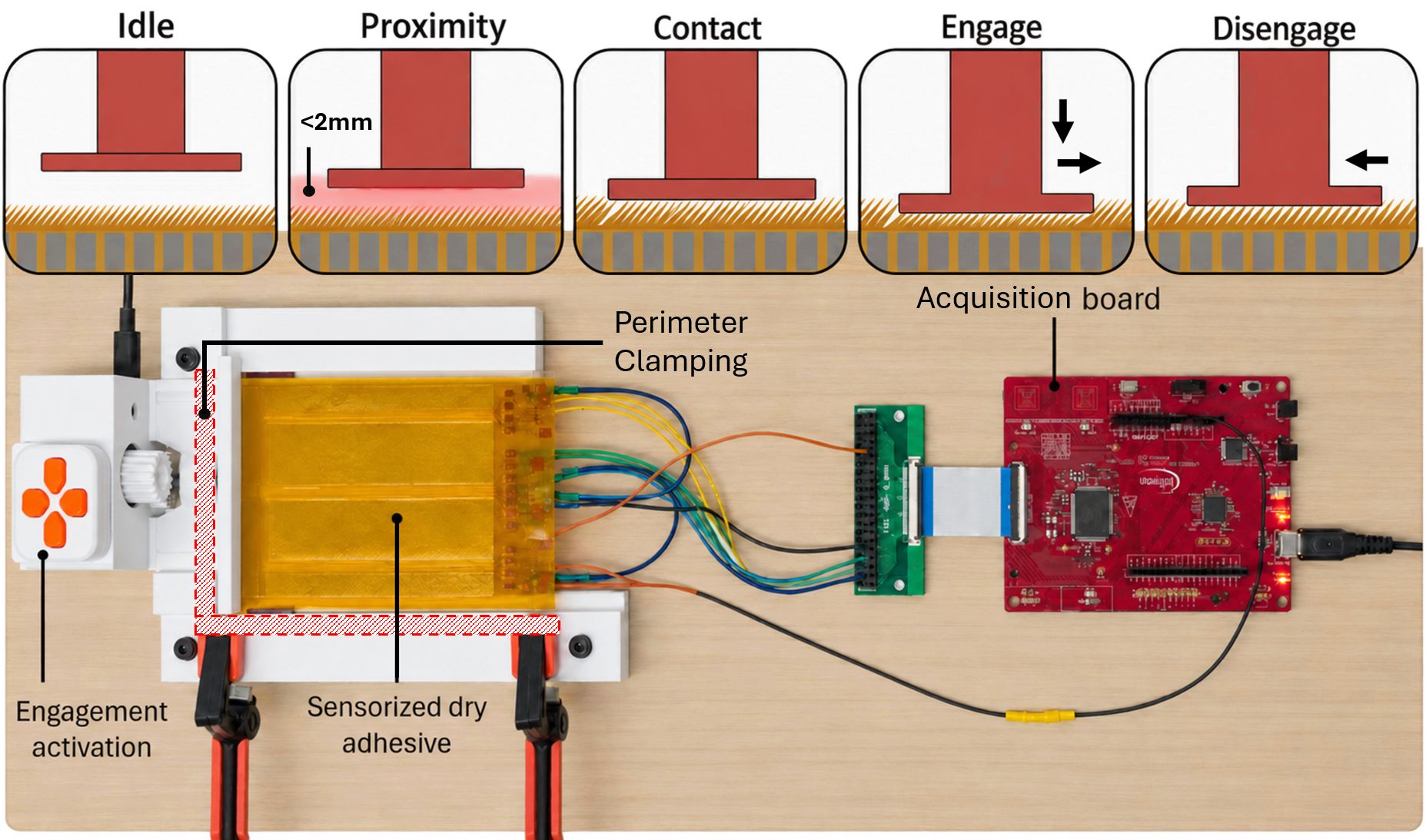}
  \caption{Experimental setup. A custom sliding mechanism integrated into a 3D-printed fixture provides controlled engagement to improve motion reliability and disengagement of the sensorized gecko-inspired adhesive. The renders illustrate the five interaction states considered in this work.}
  \label{fig:setup}
\end{figure}

An experimental platform was developed to evaluate the feasibility of real-time interaction-state recognition using a sensorized gecko-inspired adhesive interface. The adhesive material consists of a wedge-type dry adhesive developed by the Biomimetics and Dexterous Manipulation Laboratory (BDML) at Stanford University \cite{Emj2025}. A schematic of the experimental setup is shown in Fig.~\ref{fig:setup}.

Capacitance measurements were acquired using the CAPSENSE PSoC 4100S Max Pioneer Kit, which integrates fifth-generation CAPSENSE technology and provides high signal-to-noise ratio together with improved environmental compensation. The adhesive patch was mounted via $2/4$-perimeter clamping on a custom-designed fixture that securely supported the sample while enabling controlled engagement with a target surface (See Fig.  \ref{fig:setup}). During operation, the adhesive remained stationary, whereas the target was translated parallel to the adhesive interface to generate the shear displacement required for wedge activation.

The sensing electrodes embedded within the adhesive were connected directly to the Programmable System-on-Chip (PSoC) acquisition board for continuous real-time capacitance measurement. Preliminary sensor characterization identified one sensing taxel with the highest signal stability and the lowest cross-coupling between neighbouring electrodes. This sensing channel was used throughout the experiments for data acquisition to maximize measurement consistency. Sensor acquisition parameters, including sensitivity and filtering, were experimentally tuned to ensure stable and repeatable measurements.

To improve repeatability and reduce disturbances introduced by manual operation, the engagement process was automated using an ESP32 microcontroller driving a miniature stepper motor. This mechanism generated the translational motion required for adhesive engagement while minimizing parasitic capacitance variations introduced by hand proximity. The target surface consisted of a glass block (25 mm $\times$ 76 mm, 6 mm). Glass was selected because it provides a smooth, repeatable surface that is well suited for characterizing gecko-inspired adhesive performance under controlled laboratory conditions.

\subsection{Data Acquisition and Dataset Generation}

Capacitance measurements were continuously acquired from the sensorized gecko-inspired adhesive interface at a sampling rate of 20~Hz and transmitted to a host computer (Dell Latitude 5400, Intel Core i5 8365U, 16Gb RAM) via a UART serial connection. Before each acquisition session, a baseline calibration was performed to establish the reference capacitance of the sensor. During extended recording sessions, the baseline was periodically updated to compensate for environmental variations, such as temperature drift. The recorded signals were subsequently processed following the pre-processing pipeline described in Section~III.


To facilitate reliable data annotation, a custom web-based acquisition interface was developed. Rather than continuously recording complete interaction sequences, the interface allowed individual interaction states to be recorded separately. This strategy reduced labelling uncertainty by ensuring that each recorded segment corresponded to a well-defined quasi-static physical configuration.

For each interaction state, capacitance measurements were recorded for approximately 5--15~s while maintaining a constant physical configuration. Data acquisition was manually controlled: recording was initiated once the desired interaction state had been established, paused while the adhesive was reconfigured, and resumed for the subsequent state. Approximately twenty independent acquisition sessions were performed under slightly varying operating conditions to increase dataset diversity and improve robustness to experimental variability.



Five interaction states were considered:

\begin{itemize}
    \item \textbf{\textit{Idle}}: No object present near the adhesive interface.
    \item \textbf{\textit{Proximity}}: Target positioned $\approx$~1.5--2~mm above the adhesive surface using calibrated 3D-printed spacers.
    \item \textbf{\textit{Contact}}: Direct contact between the target and the adhesive without shear activation.
    \item \textbf{\textit{Engaged}}: Adhesive activated through the combined application of normal preload and shear displacement.
    \item \textbf{\textit{Disengaged}}: Adhesive released following shear removal and normal preload after engagement.
\end{itemize}

The resulting dataset comprised labelled steady-state recordings for all five interaction classes. Following the augmentation procedure described in Section~III, an expanded training dataset was generated to improve model robustness. Table~\ref{tab:data_summary} summarises the final dataset used for classifier training and evaluation.

\begin{table}[h]
\centering
\vspace*{0.25cm}
\caption{Summary of the steady-state and augmented datasets used for classifier training and evaluation.}
\begin{tabular}{lcc}
\toprule
\textbf{Interaction State} & \textbf{Steady Data} & \textbf{Augmented Data} \\
\toprule
Idle&4,167&15,298\\
Proximity&7,948&13,428\\
Contact&3,976&6,903\\
Engaged&3,704&6,771\\
Disengaged&3,676&5,991\\
\midrule
\textbf{Total}&\textbf{23,471}&\textbf{48,391}\\
\bottomrule
\end{tabular}
\label{tab:data_summary}
\end{table}

\subsection{Offline Classification Performance}

The proposed interaction-state recognition framework was first evaluated offline using the steady-state and augmented datasets described in the previous subsection. Six classical machine learning classifiers were trained using the extracted feature vectors and evaluated on a held-out test set. Table~\ref{tab:classifier_performance} summarizes the resulting classification accuracies.

 \begin{table}[!h]
\centering
\caption{Classification accuracy on steady-state and augmented datasets.}
\resizebox{\columnwidth}{!}{%
\begin{tabular}{lcc}
\toprule
\textbf{Classifier} & \textbf{Steady Data} & \textbf{Augmented Data} \\
\midrule
Gradient Boosting & 0.9790 & 0.9439 \\
Logistic Regression & 0.7917 & 0.9168 \\
K-Nearest Neighbors & 0.9757 & 0.9153 \\
Support Vector Machine & 0.8417 & 0.9134 \\
Random Forest & 0.9764& \textbf{0.9565} \\
Extremely Randomized Trees & \textbf{0.9816} & 0.9421 \\
\bottomrule
\end{tabular}%
}
\label{tab:classifier_performance}
\end{table}

All evaluated classifiers achieved high accuracy on the steady-state dataset, indicating that the extracted features effectively characterize the five interaction states under controlled conditions. The Extremely Randomized Trees classifier obtained the highest accuracy (98.16\%), closely followed by Random Forest (97.64\%).

A different trend is observed for the augmented dataset, which better captures the variability expected during practical operation. Under these conditions, Random Forest~(RF) achieved the highest classification accuracy (95.65\%), outperforming the other models while maintaining stable performance across all interaction classes. While steady-state data provide a cleaner classification setting, evaluation on augmented data reduces the risk of overestimating performance under idealized conditions and provides a more realistic assessment of model robustness to signal variability and dynamic transitions. Based on these results, RF was selected for subsequent online evaluation. Table~\ref{tab:rf_hyperparameters} reports the optimized hyperparameters of the RF obtained using Optuna.
 
\begin{table}[h]
\centering

\caption{Optimized Random Forest hyperparameters obtained through Optuna search}
\label{tab:rf_hyperparameters}
\begin{tabular}{lc}
\hline
\textbf{Hyperparameter} & \textbf{Value} \\
\hline
Number of estimators ($n_{\mathrm{estimators}}$) & 550 \\
Maximum depth ($max_{\mathrm{depth}}$) & None \\
Minimum samples split ($min_{\mathrm{samples\_split}}$) & 6 \\
Minimum samples leaf ($min_{\mathrm{samples\_leaf}}$) & 2 \\
Maximum features ($max_{\mathrm{features}}$) & None \\
Bootstrap sampling ($bootstrap$) & True \\
\hline
\end{tabular}
\end{table}

\begin{figure}[!h]
   \centering
   \includegraphics[width=1.0\columnwidth]{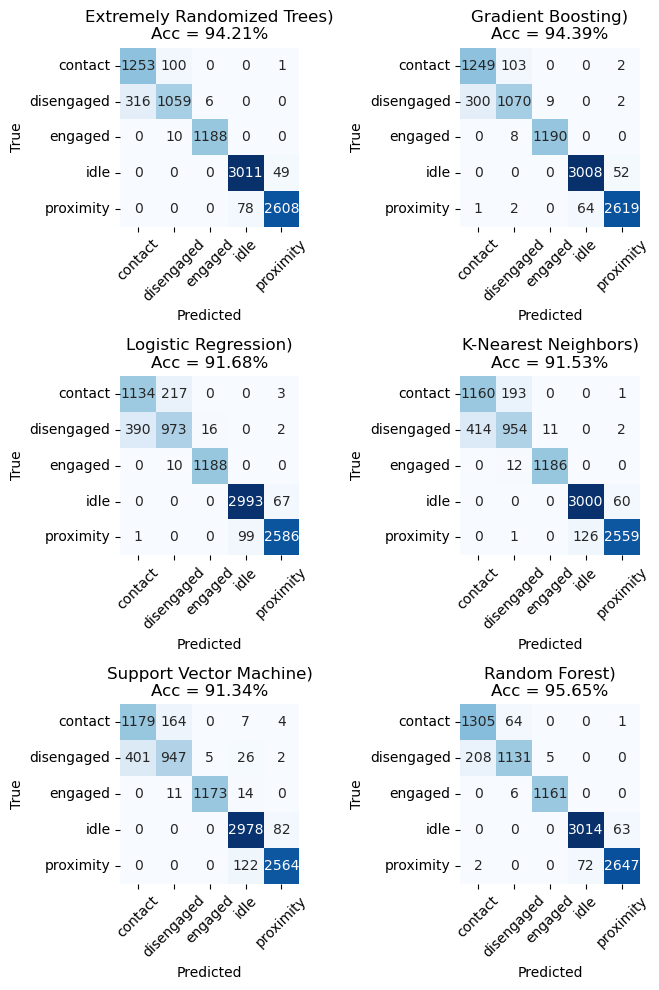}
   \caption{Confusion matrices on the augmented dataset. The largest confusion occurs between the \textit{Contact} and \textit{Disengaged} states due to their similar mechanical configurations.}
   \label{fig:conf_matrix_classifiers}
 \end{figure}

To better understand the behaviour of the different classifiers, Fig.~\ref{fig:conf_matrix_classifiers} presents the corresponding confusion matrices. States associated with clearly distinct physical configurations, particularly \textit{Idle} and \textit{Engaged}, are consistently classified with high accuracy across all models due to their distinct capacitance signature. In contrast, the greatest ambiguity occurs between the \textit{Contact} and \textit{Disengaged} states. These two classes correspond to mechanically similar configurations, differing primarily in the residual deformation of the adhesive wedges immediately following disengagement, which produces partially overlapping capacitance signatures, leading to occasional misclassification. Functionally, distinguishing these states is critical because Contact occurs during initial approach prior to engagement, whereas Disengaged follows shear release during detachment.

The results indicate that steady-state interaction-state classification is feasible using embedded capacitive sensing and lightweight machine learning. At the same time, the observed confusion between mechanically similar states suggests that static feature representations alone are insufficient to fully characterize the dynamics of adhesive interaction, motivating the transition-aware framework presented in the following subsection.

\subsection{Online Inference Performance}
\begin{figure}[!h]
  \centering
  \includegraphics[width=1\columnwidth]{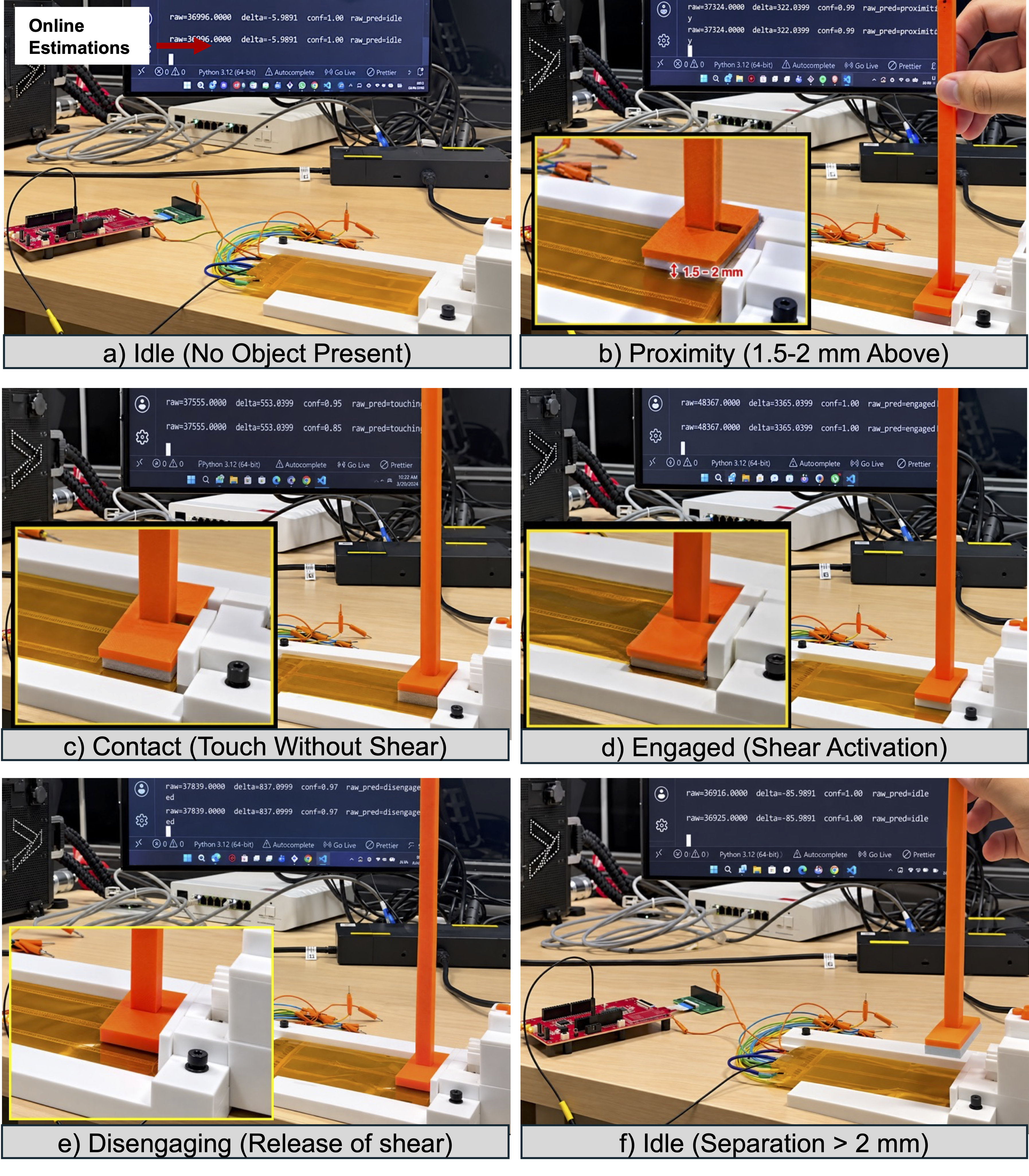}
  \caption{Representative images of the online evaluation covering the five interaction states.}
  \label{fig:class_desc}
\end{figure}

The Random Forest (RF) classifier, which achieved the best offline performance on the augmented dataset, was selected for real-time evaluation. The objective of this experiment was to assess the complete interaction-state recognition framework under live operating conditions, including signal acquisition, preprocessing, feature extraction, classification, and state-machine post-processing. Prior to each experiment, a 20~s baseline recording was acquired to establish the reference capacitance used for signal normalization. The adhesive interface was then sequentially driven through the five interaction states shown in Fig.~\ref{fig:class_desc}. During the experiment, capacitance measurements were continuously processed by the complete inference pipeline, producing an interaction-state estimate at each sampling instant.

\begin{figure*}[!h]
  \centering
  \includegraphics[width=0.98\textwidth]{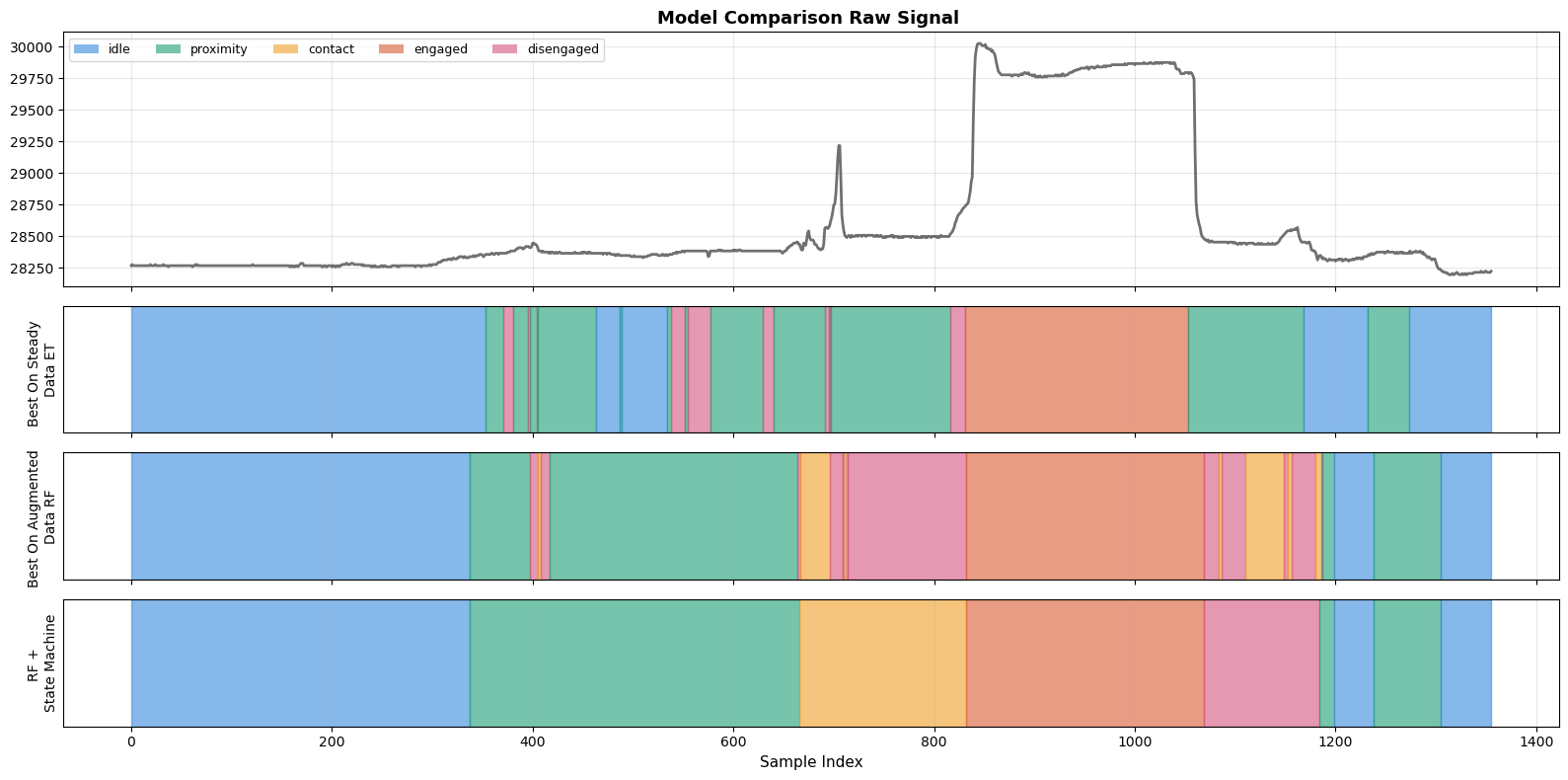}
  \caption{
  Comparison of interaction-state predictions during online operation, where colors represent the different interaction states. The top plot shows the measured capacitance signal, while the three remaining plots present the predicted interaction states obtained using (top) ET trained on the steady-state dataset, (middle) Random Forest trained on the augmented dataset, and (bottom) Random Forest combined with the proposed state-machine post-processing (RF+SM).}
  \label{fig:comp_ml_live}
\end{figure*}

The complete framework, including signal acquisition, preprocessing, feature extraction, Random Forest inference, and state-machine post-processing, achieved an average processing time of 53~ms per inference cycle. At the adopted sampling rate of 20~Hz, corresponding to a 50~ms sampling interval, the complete processing time is approximately one sampling period. This processing time demonstrates that the proposed framework is suitable for real-time embedded implementation while maintaining continuous interaction-state estimation.

To evaluate the proposed framework, three models were compared during online operation: (i) the Extremely Randomized Trees (ET) classifier trained on the steady-state dataset, (ii) the Random Forest (RF) classifier trained on the augmented dataset, and (iii) the Random Forest classifier combined with the proposed state machine (RF+SM). Figure~\ref{fig:comp_ml_live} compares the interaction-state sequences produced by each approach.

The ET classifier produces highly fragmented predictions and frequent spurious state transitions, indicating limited robustness when deployed under dynamic operating conditions. Training with the augmented dataset substantially improves temporal consistency, with the Random Forest classifier generating more coherent interaction-state sequences. Nevertheless, occasional misclassifications remain, particularly during transitions between mechanically similar states such as \textit{Contact} and \textit{Disengaged}.

The proposed RF+SM framework produces the most stable and physically consistent interaction-state estimates. By combining data-driven classification with the temporal filtering and transition constraints described in Section~III, isolated prediction errors are effectively suppressed while physically implausible state transitions are eliminated. As illustrated in Fig.~\ref{fig:comp_ml_live}, the resulting interaction-state sequence closely follows the expected progression of the adhesive interaction and exhibits significantly fewer spurious transitions than the standalone classifiers.

\begin{table}[!h]
\centering
\caption{Performance comparison of the evaluated models during online operation. (W) denotes weighted metrics.}
\resizebox{\columnwidth}{!}{%
\begin{tabular}{lccccc}
\toprule
\textbf{Model} & \textbf{Accuracy} & \textbf{Precision (W)} & \textbf{Recall (W)} & \textbf{F1 (W)} & \textbf{F1 (Macro)} \\
\midrule
ET & 0.6313	&0.5427&	0.6313	&0.5784&	0.4598\\
RF & 0.7802&0.7942&0.7802&0.7787&0.6417 \\
RF+SM & \textbf{0.9314} & \textbf{0.9364} & \textbf{0.9314} & \textbf{0.9307} & \textbf{0.9343} \\
\bottomrule
\end{tabular}%
}
\label{tab:model_performance_online}
\end{table}

\begin{figure*}[!h]
  \centering
  \includegraphics[width=0.85\textwidth]{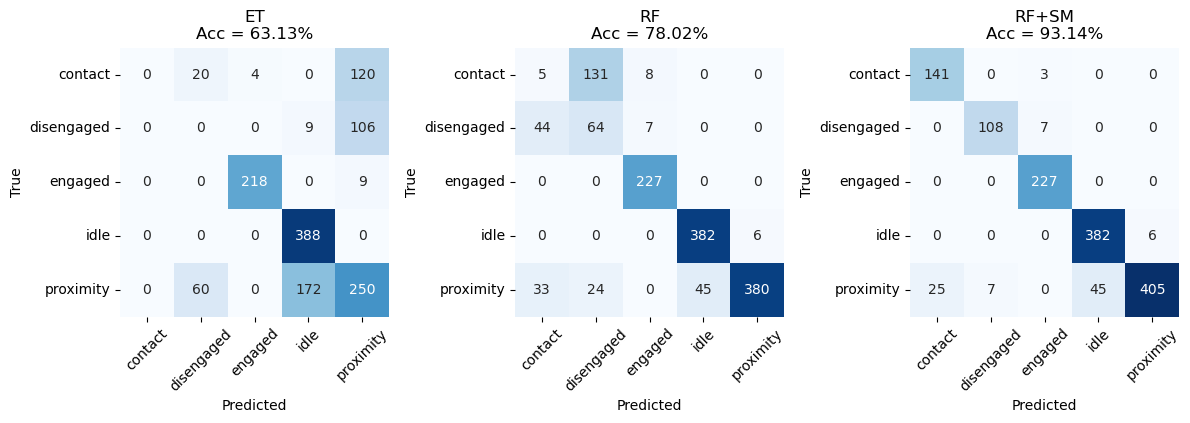}
  \caption{Confusion matrices for the online evaluation.}
  \label{fig:cm_live}
\end{figure*}

These qualitative observations are supported by the quantitative results summarized in Table~\ref{tab:model_performance_online}. Compared with the standalone Random Forest classifier, the proposed RF+SM framework improves the online classification accuracy from 78.02\% to 93.14\%, while increasing the macro F1-score from 0.6417 to 0.9343. This demonstrates that incorporating simple physics-informed constraints substantially improves the robustness of online interaction-state recognition without increasing the computational complexity of the underlying machine learning model.

Figure~\ref{fig:cm_live} provides further insight into the behaviour of the three models through their confusion matrices. The ET classifier exhibits substantial confusion between the \textit{Contact} and \textit{Disengaged} states, indicating that training on steady-state data alone is insufficient to capture the variability encountered during dynamic operation. The Random Forest classifier improves class separation but still struggles to distinguish transient interaction phases with similar capacitance signatures. Incorporating the state machine substantially reduces these ambiguities by enforcing temporally consistent and physically admissible state transitions, resulting in the most balanced per-class performance across all five interaction states.

\section{Conclusions}

This paper presented a framework for real-time interaction-state recognition in a sensorized gecko-inspired dry adhesive using embedded capacitive sensing and lightweight machine learning. The proposed approach estimates five interaction states throughout the adhesive lifecycle, providing the adhesive interface with awareness of its physical interaction with the environment. Experimental results demonstrated that classical machine learning methods can reliably classify interaction states from capacitive measurements, with Random Forest achieving the best offline performance. Furthermore, incorporating a lightweight rule-based state machine significantly improved the robustness of online inference by enforcing temporally consistent and physically admissible state transitions, increasing the online classification accuracy from 78.02\% to 93.14\% while maintaining a processing time of 53~ms per inference cycle. Although this study was conducted under controlled laboratory conditions, the results demonstrate the feasibility of interaction-aware gecko-inspired adhesive interfaces as a first step toward autonomous contact monitoring for space robotic applications. While conductive spacecraft parts  yields higher capacitance shifts ($\Delta C$) that simplify classification, surface roughness introduces micro-air gaps that alter baseline dielectric coupling. Future work will focus on extending the experimental validation to a wider range of contact surfaces, loading conditions, and dynamic interaction scenarios.

\addtolength{\textheight}{-12cm}   







\section*{ACKNOWLEDGMENT}
The authors would like to express their gratitude to the Biomimetics and Dexterous Manipulation Laboratory (BDML) at Stanford University for providing the sensorized gecko-inspired adhesive material and for their support in setting up the system.


\bibliography{references}
\bibliographystyle{IEEEtran}

\end{document}